\documentclass[letterpaper]{article} 
\usepackage{aaai24}  
\nocopyright
\usepackage{times}  
\usepackage{helvet}  
\usepackage{courier}  
\usepackage[hyphens]{url}  
\usepackage{graphicx} 
\usepackage{natbib}  
\usepackage{caption} 
\usepackage{amsmath}
\usepackage{float}

\usepackage{algorithm}
\usepackage{algorithmic}

\usepackage{newfloat}
\usepackage{listings}
\DeclareCaptionStyle{ruled}{labelfont=normalfont,labelsep=colon,strut=off} 
\floatstyle{ruled}
\newfloat{listing}{tb}{lst}{}
\floatname{listing}{Listing}
\usepackage{listings}
\usepackage{color}
\usepackage{multirow}
\usepackage{booktabs}
\usepackage{graphicx} 
\usepackage{pifont}

\title{CGMI: Configurable General Multi-Agent Interaction Framework}
\author{
    Jinxin Shi\textsuperscript{\rm 1},
    Jiabao Zhao\textsuperscript{\rm 1*},
    Yilei Wang\textsuperscript{\rm 1},
    Xingjiao Wu\textsuperscript{\rm 2},
    Jiawen Li\textsuperscript{\rm 1},
    Liang He\textsuperscript{\rm 1}
}
\affiliations{

    \textsuperscript{\rm 1}School of Computer Science and Technology, East China Normal University, Shanghai, China\\
    \textsuperscript{\rm 2}School of Computer Science, Fudan University, Shanghai, China\\
    52275901016@stu.ecnu.edu.cn, 
    jbzhao@mail.ecnu.edu.cn,
    wangyilei@mail.ecnu.edu.cn,
    xjwu\_cs@fudan.edu.cn,
    52275901026@stu.ecnu.edu.cn,
    lhe@cs.ecnu.edu.cn
}

\usepackage{bibentry}

\begin{document}

\maketitle

\begin{abstract}
Benefiting from the powerful capabilities of large language models (LLMs), agents based on LLMs have shown the potential to address domain-specific tasks and emulate human behaviors. However, the content generated by these agents remains somewhat superficial, owing to their limited domain expertise and the absence of an effective cognitive architecture. To address this, we present the Configurable General Multi-Agent Interaction (CGMI) framework, designed to replicate human interactions in real-world scenarios. Specifically, we propose a tree-structured methodology for the assignment, detection, and maintenance of agent personality. Additionally, we designed a cognitive architecture equipped with a skill library based on the ACT* model, which contains memory, reflection, and planning modules. We have also integrated general agents to augment the virtual environment’s realism. Using the CGMI framework, we simulated numerous classroom interactions between teacher and students. The experiments indicate that aspects such as the teaching methodology, curriculum, and student performance closely mirror real classroom settings. We will open source our work.
\end{abstract}

\section{Introduction}
\label{sec:int}
Agent-based social simulation (ABSS) simulates social interactions in a virtual environment. By observing agent behavior, we can explore complex social phenomena and verify the effects of different social strategies in a controlled setting\cite{davidsson2002agent}. However, improving simulation accuracy and designing complex agents remain key challenges\cite{pmlr-v202-aher23a}. With the capabilities of large language models (LLMs) such as GPT4 \cite{openai2023gpt4}, we can construct more complex environment and create more realistic agents to simulate social phenomena. However, when using LLMs to complete ABSS tasks, the following issues need to be addressed: (1) How to trigger the capabilities of LLMs to solve complex problems? (2) How to ensure that agents have a stable role and behavior output based on LLMs without forgetting? (3) How to design a communication mechanism for LLMs-based agents to truly simulate interactions?

Existing LLMs-based agents are mainly divided into action agents \cite{yao2023react, press2023measuring} and plan-and-execute agents \cite{wang2023planandsolve}. Action agents make decisions based on previous outputs and are suitable for small tasks. Plan-and-execute agents formulate and execute action plans, suitable for long-term goal tasks. However, in complex scenarios, LLMs-based agents may produce mechanical and superficial content or not execute according to the plan. Inspired by the Adaptive Control of Thought (ACT*) model \cite{anderson1983spreading}, we designed a cognitive architecture equipped with skill library for agents. Specifically, we employ the Chain of Thought (CoT) and Chain of Action (CoA) methods to extract declarative and procedural memories from the agent's working memory. During the reflection and planning processes, content is retrieved from the skill library, ensuring deeper and more specialized insights.

Assigning each intelligent agent with a unique identity, personality, and capability \cite{wang2023unleashing} can offer a more humanized and emotional interactive experience, and also enhance the realism of simulating complex social scenarios \cite{argyle2023out}. Although LLMs like GPT4 possess strong role-playing capabilities, we found that LLMs tend to forget the original character settings in multi-turn dialogues and make decisions that are inconsistent with the character's design. Additionally, due to the limitations of the context window, it's challenging to set roles comprehensively and in fine detail. To address these issues, this paper introduces a tree-structured persona model for character assignment, detection, and maintenance, which is beneficial for agent interaction performance.

Influenced by assistant repeats instruction, infinite loop of messages, and conversation termination conditions, it remains challenging for chat agents to automatically collaborate to accomplish tasks in specific scenarios\cite{li2023camel}. Setting scenario-adapted general agents is used to solve scenario-specific tasks for role agents, can help role agents avoid the aforementioned problems and enhance the realism of virtual scenes. For this purpose, this paper explores a Configurable General Multi-Agent Interaction Framework (CGMI), that can simulate real-life scenarios by binding general agents with role agents.

In this work, we take the "classroom teaching scenario" as an example, employing the CGMI framework to simulate the teaching process between "teacher" and "students", including teacher agent, student agents, assistant agents and supervisory agents. The experimental results indicate that the  interactions in the virtual classroom aligns with actual teaching. It helps to assist in teacher instruction, evaluate teaching competencies, and validate teaching hypotheses.

In summary, the major contributions of this paper are threefold:
\begin{itemize}
\item The introduction of cognitive structure equipped with skill library, combining human cognition and skill library retrieval, enabling agents to engage in deep reflection and planning.
\item Designed a tree-structured approach for assigning, detecting, and maintaining the personal traits of agents, which reduces memory pressure on agents and improves stability.
\item The construction of a Configurable General Multi-agent Interaction framework (CGMI), supporting social experimental research in specific scenarios.
\end{itemize}
\section{Related Work}

In this section, we will review agent research for solving domain problems, as well as agent research for simulating real human interaction processes.
\subsection{Agents for Solving Domain Problems}
Recent studies in LLMs have explored the utilization of agent systems for domain-specific tasks across various sectors. In healthcare, \cite{nair2023dera} introduced a multi-agent system that enhances treatment recommendations via communication feedback. \cite{qian2023communicative} presented CHATDEV: a simulated development team where agents oversee design, coding, testing, and documentation, thereby ensuring effective game development coordination. \cite{alexandru2015enhanced} designed a multi-agent e-learning environment tailored for education, providing customized support for instructional decisions. ChemCrow, highlighted in \cite{bran2023chemcrow}, formulated a framework that grants agents access to external knowledge repositories, consequently amplifying their efficacy in areas like organic synthesis, drug discovery, and materials design. \cite{wang2023describe} unveiled the DEPS interactive planning technique, addressing long-term planning challenges within the Minecraft game. Collectively, these investigations illuminate agent applications tailored to particular domains and hurdles.

\subsection{Agents for Simulating Human Interactions}
A subsequent line of research focuses on crafting agents that emulate human social behaviors. \cite{park2022social} fashioned a multi-agent town emulating authentic human activities, including orchestrating social parties. \cite{li2023camel} delved into an agent communication framework that facilitates varied social roles and simulates AI social patterns. Emphasizing the importance of social situational learning, \cite{krishna2022socially} developed an interactive agent capable of querying individuals online to assimilate visual knowledge. In the educational realm, \cite{markel2023gpteach} employed GPT and other LLMs to mimic students, thus offering tangible training avenues for educators. \cite{jiang2023personallm} explored the simulation of consistent personality and gender variations using conditional language models. Cumulatively, these studies accentuate agents' capacities to assimilate or mimic human social interactions.

\begin{figure}[!hbtp]
\includegraphics[width=\linewidth]{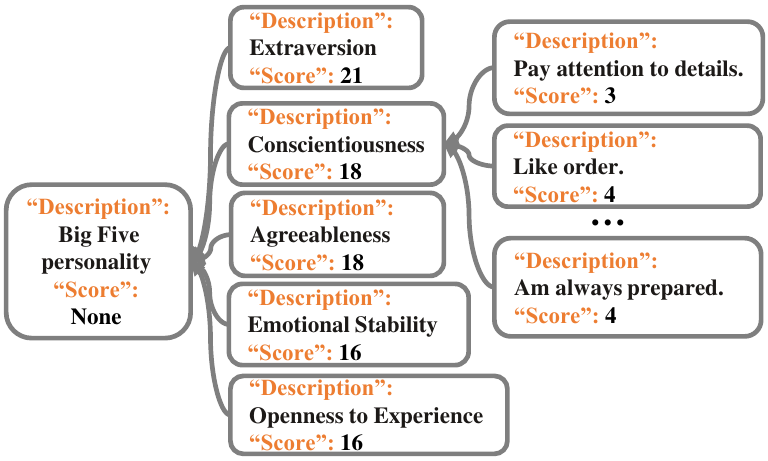}
\caption{Tree structure of the Big Five Personality Scale. The root node has five sub-nodes, representing five coarse personalities. Their dimension values range from 5-25, and each coarse personality has five fine-grained leaf nodes, with dimension values ranging from 1-5. The larger the value, the more pronounced the characteristics of agents.} 
\label{PM}
\end{figure}

\begin{figure*}[t]
\centering
\includegraphics[width=\linewidth]{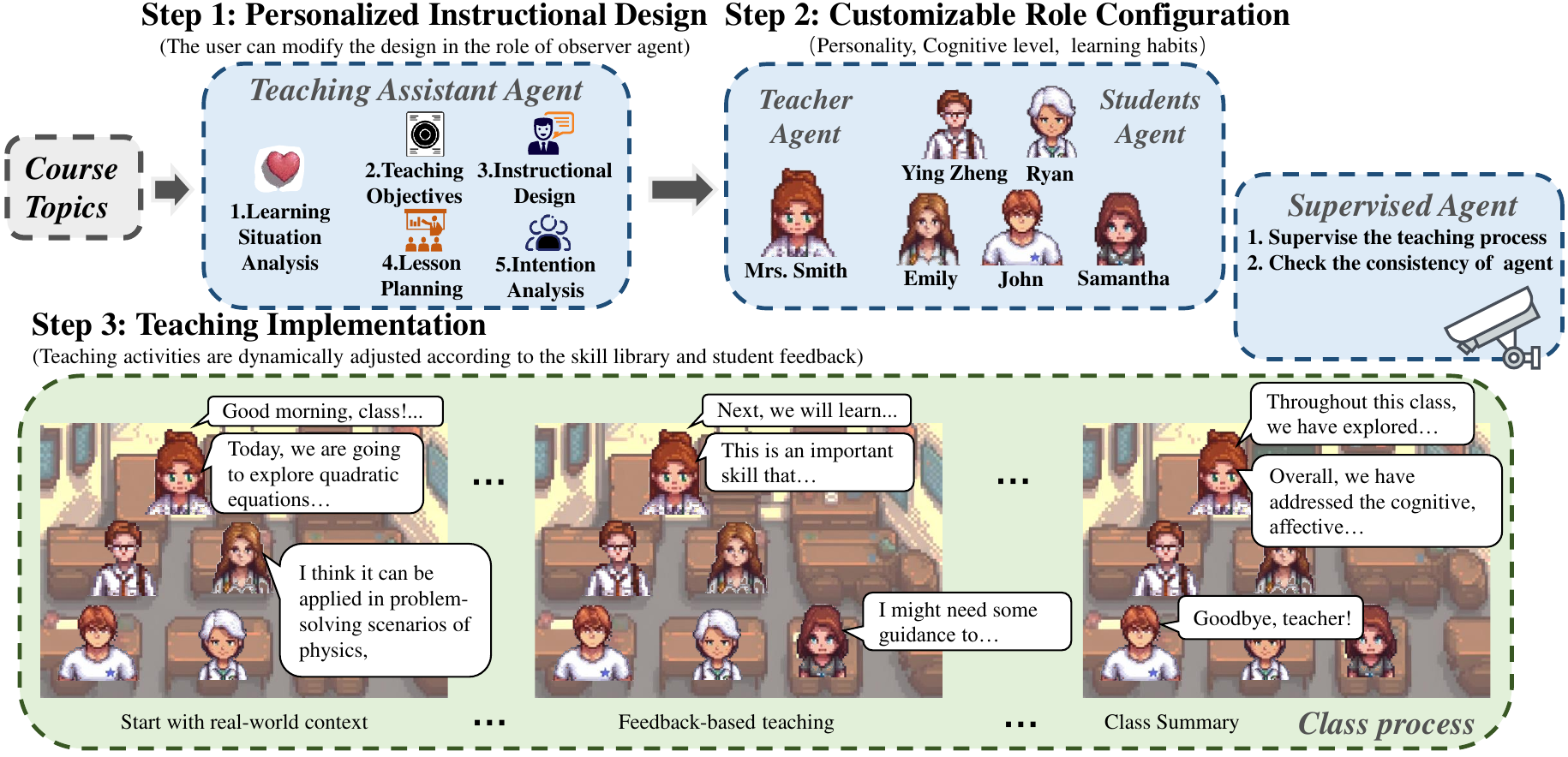} 
\caption{Based on CGMI, a classroom teaching scenario is constructed. This scenario includes 3 general intelligent agents (teaching assistant agent, teaching process supervisor agent, consistency checker agent) and 6 role agents (teacher Mrs. Smith, student Ying Zheng, student Emily, student John, student Ryan and student Samantha). After the user inputs the course topic, the virtual classroom teaching scenario launches. The teaching assistant agent generates corresponding teaching plans and distributes them to Mrs. Smith and the teaching process supervisor agent. Mrs. Smith divides the teaching process into stages according to the plan. The teaching process supervisor agent monitors whether the current stage has ended and decide whether to enter the next stage. Before each role agent's statement, the consistency checker agent detects and maintains consistency between its personality and statement content. When Mrs. Smith asks the class questions, the consistency checker agent judges each student's willingness to answer based on personality and classroom status, simulating real hand-raising.}
\label{fig2}
\end{figure*}

\section{Method}
\label{sec:met}

In this section, the tree-structured approach for personality assignment, detection and maintenance, the cognitive structure model enhanced with a skill library, and the construction process of CGMI will be introduced respectively. As shown in Figure 2, the process of reconstructing the "classroom teaching" scenario based on CGMI is displayed.

\subsection{Tree-Structured Persona Model}
Agent entities with unique personalities can not only complete specific tasks, but also enhance the authenticity of interactions \cite{qian2018assigning, PUDANE2017517}. In addition to setting specific personalities for agent entities, it is also necessary to set related styles according to the application scenario. For example, in teaching, teacher and students can have their own teaching and learning styles. However, if only a rough persona is set for agents, the personalized differences in its interactions are not obvious, and its stability will decrease as the complexity of roles, scenarios, and the length of the context increase \cite{jiang2023personallm}.

To solve this problem, this work proposes a tree-structured persona model for personality assignment, detection, and maintenance. We referred to the Big Five Personality Scale \cite{john1999big}, the teaching style scale \cite{grigorenko1993thinking}, and the learning style scale \cite{soloman2005index}, and designed a tree structure to help agents remember and set different personas. Taking personality setting as an example, as shwon in Figure \ref{PM}, we built a personality scale $T=\{N_1, N_2, ..., N_n\}$ based on the Big Five Personality Scale, where $n=26$. $N_1$ is the root node, and $N_2$ to $N_n$ are child nodes. Each node Ni includes a description $D_i$ and a score $S_i$. As shown in Algorithm \ref{alg:algorithm}, we use depth-first traversal to set personality traits for the intelligent entity $A$.

\begin{algorithm}[t]
\caption{The process of endowing the Big Five personalities through Deep First Traverse (DFS) implementation.}
\label{alg:algorithm}
\textbf{Input}: Big Five Scale $T$, Agent $A$\\
\textbf{Output}: $A=\{T\}$
\begin{algorithmic}[1] 
\STATE Define stack 
\STATE Push root node of $T$ into stack
\WHILE{stack is not empty}
\STATE $N_i$ = stack.pop()
\STATE $A$ get($N_i.D_i, N_i.S_i$)
\IF {$N_i$ has child nodes}
\STATE Push all child nodes of $N_i$ into stack
\ENDIF
\ENDWHILE
\STATE \textbf{return} $A=\{T\}$
\end{algorithmic}
\end{algorithm}

During the detection and maintenance process, this paper adopts an efficient random testing method, with the following specific steps: (1) Randomly select $m$ coarse-grained personalities for testing; (2) If the test is correct, select $m$ fine-grained personalities under these $m$ coarse-grained personalities for further testing. If the fine-grained test is also correct, it is believed that the agent's personality memory is complete; (3) If an error occurs at any stage, the real values of all selected personalities will be informed to the agent to restore its personality memory.

This random testing method is not only efficient and comprehensive but also saves contextual window resources. Multi-level testing can avoid the illusion of unchanged coarse-grained personality due to changes in fine-grained personality. This method can also be applied to other related character scales, as detailed in Appendix.

\subsection{Cognitive architecture equipped with skill library}

Over time, as interactions between the agent and its environment accumulate, there's a marked increase in the volume and intricacy of the agent's memory stream.\cite{park2023generative,weng2023prompt} This proliferation necessitates an advanced cognitive architecture to process the burgeoning data. However, the current cognitive architecture embedded in LLMs-based agents can only allow agents to plan and reflect in a linear fashion, reminiscent of an assembly line. To redress this shortfall, this paper introduces the cognitive architecture infused with a domain-specific skill library, rooted in the Adaptive Control of Thought (ACT*) paradigm\cite{anderson1983spreading}. This novel architecture facilitates parallel and bidirectional planning and reflection, drawing upon the agent's memory and skill repository, thus steering agent development towards enhanced adaptive control and rational deliberation akin to human cognition.

\begin{figure}[t]
\includegraphics[width=\linewidth]{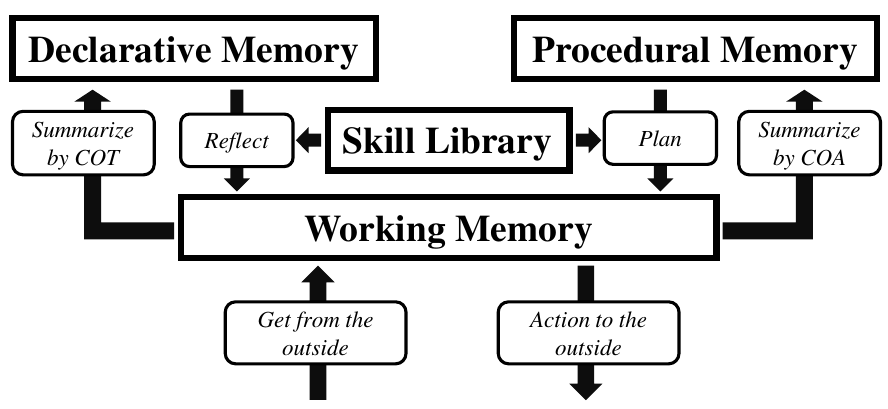}
\caption{The cognitive architecture with skill library.} 
\label{fig3}
\end{figure}

Central to this cognitive framework are four pivotal components, as delineated in Figure 3. The foundational pillars of agent cognition are Declarative $(M_d)$ and Procedural Memory $(M_p)$. The former embodies the agent's library of factual knowledge, encompassing data on objects, individuals, locales, occurrences and their interconnections, serving as the cornerstone for rational deduction. Procedural memory, on the other hand, comprises operational guidelines that empower the agent to pursue objectives and surmount challenges. These guidelines operate by matching with facts stored declaratively, triggering actions geared towards achieving specific objectives. Skill Library ($L$) is a configurable domain knowledge base that provides domain knowledge for the reflective planning of intelligent agents. It can be viewed as a compilation of the agent's abilities to leverage its knowledge in situation-specific ways. Working Memory $(M_w)$ is an agile, self-refreshing module acting as a bridge between memory and the external milieu. It not only directs agent actions based on processed memories but also assimilates external data, subsequently refining it into declarative and procedural knowledge via the Chain of Thoughts (CoT) and Chain of Actions (CoA).

When starting interaction, an agent, denoted as $A=\{T, B\}$ and equipped with the cognitive architecture $B=\{M_w, M_d, M_p, L\}$, seamlessly activates these four components, ensuring prolonged engagements in multifaceted settings. Formally, the mechanism through which the agent gleans information from the external realm at a given time $t$ is depicted as $F_{get}(t)$.

Upon temporary storage in $M_w$, the agent $A$ distills this information using thought and action chains, leading to the formation of Declarative and Procedural Memory:
\begin{gather}
M_d(t) = F_{sum}(P_{cot} + M_w(F_{get}(t))) \\
M_p(t) = F_{sum}(P_{coa} + M_w(F_{get}(t))) 
\end{gather}
where $P_{cot}$ signifies the CoT prompt (e.g., "Summarize the class content sequentially"), while $P_{coa}$ denotes the CoA prompt (e.g., "Detail the pedagogical steps"). $F_{sum}$ delineates the process of condensing information within the Working Memory.
In subsequent interactions, when agent $A$ readies its response for moment $t+1$, it first taps into $M_d$, $M_p$, and $L$, extracting reflections and strategies from the preceding moment, $t$, which then translates into overt actions:
\begin{gather}
R(t) = F_{ref}(M_d(t) + L)\\
P(t) = F_{pla}(M_p(t) + L)\\
ACT(t+1) = F_{act}(R(t) + P(t) + M_w(F_{get}(t)) 
\end{gather}
where $F_{ref}$ and $F_{pla}$ illustrate the reflection and synthesis processes for Declarative and Procedural Memory at moment $t$, respectively. $R(t)$ and $P(t)$ represent the reflective and strategic outcomes at time t, while $F_{act}$ encapsulates the amalgamation of these insights, plans, and the skill repertoire to forge $ACT(t+1)$.

\begin{figure*}[t]
\centering
\includegraphics[width=\linewidth]{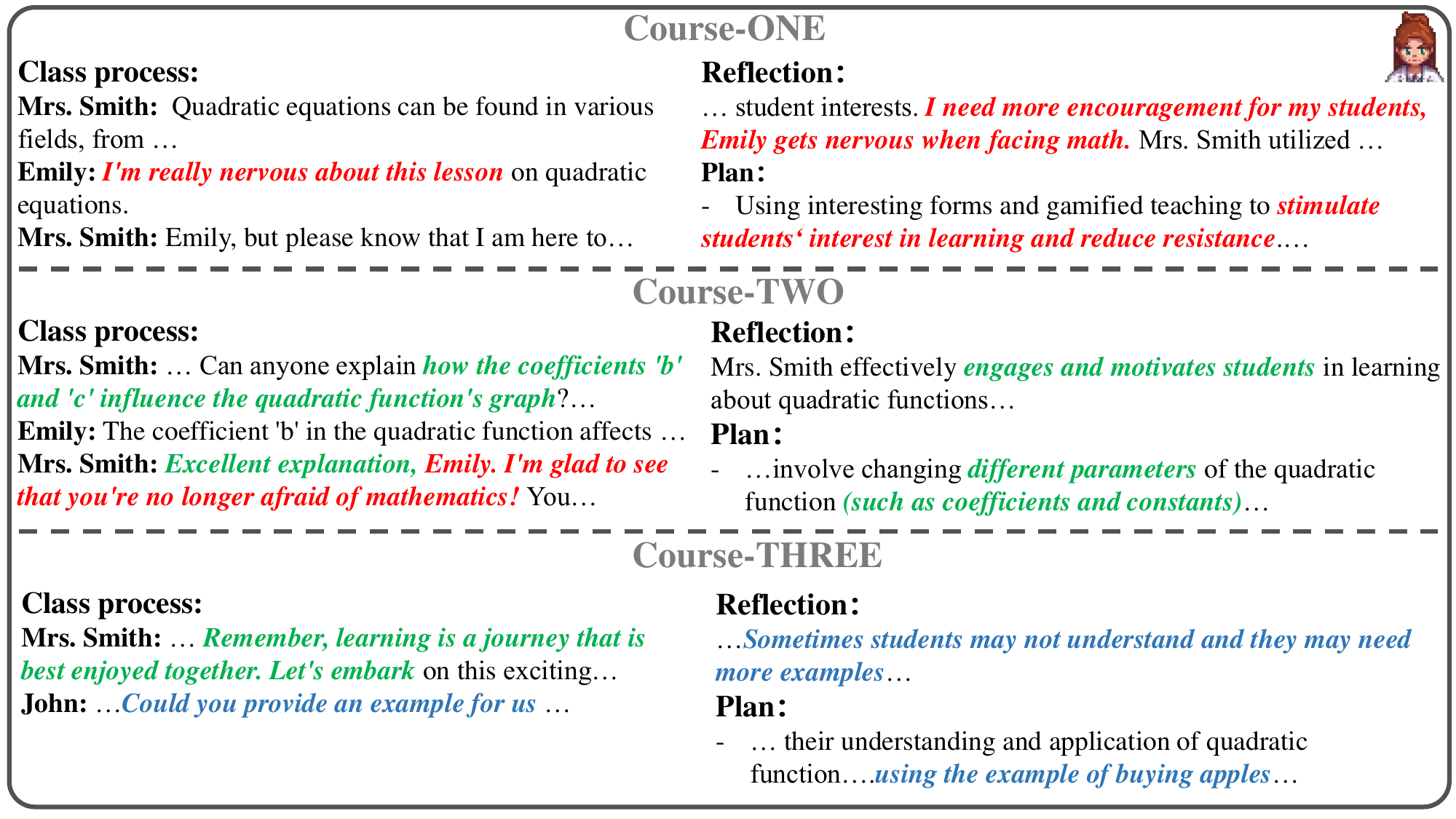}
\caption{Teacher Mrs Smith's classroom experience and her reflection and planning in virtual classroom. The red, green, and blue characters in the picture represent the events discovered by the teacher in three different classes. The teacher reflects and plans on these events, and serves as a focus in the subsequent teaching process.} 
\label{fig4}
\end{figure*}

\subsection{Configurable General Multi-Agent Interaction Framework}
With the support of structured persona models and enhanced cognitive models with skill libraries, a single agent can play multiple roles in specific scenarios to complete complex tasks. However, currently, using LLMs-based agents to achieve preset goals in specific tasks often fails to present real social interactions, because simulating social phenomena requires multiple Agents to interact and cooperate in a human-like manner. Therefore, this paper introduces the Configurable General Multi-Agent Interaction Framework (CGMI) that can simulate real interactions.

In the context of classroom teaching, this paper explores how CGMI promotes interaction and collaboration among multiple agents. In addition to virtual teacher Agent and virtual student Agents, we have also designed assistant Agents responsible for setting educational goals, planning teaching schedules, and analyzing students' willingness to speak to support teacher's teaching activities. These assistant Agents can adjust their functional configurations based on specific scenarios. To ensure the quality of the interaction process, we introduced a supervisory Agent responsible for detecting "personality forgetting", ensuring that the "teacher Agent proceeds with teaching as planned", and "determining when to end the discussion". Through the CGMI framework, each intelligent entity can engage in more in-depth personalized dialogues and task completion, collaboratively creating a realistic virtual teaching environment.

Using classroom teaching as an example, based on cognitive structure and persona models, the intelligent agent $A=\{T, B\}$ can play different roles in specific scenarios. The state of the classroom at time t is represented as:
\begin{equation}
STA(t) = I(A_{tea}, A_{stu}, t)
\end{equation}
Where $I$ represents the interaction process, $A_{tea}$ represents the teacher, and $A_{stu}$ represents a set of students, denoted as $\{A_{stu_1}, A_{stu_2}, ..., A_{stu_n}\}$. Interact represents the interaction between the teacher and students.

When the lesson begins, the supervisory Agent $A_{sup}$ receives the teaching plan $TP$ and the multi-stage teaching process $TS$ decomposed by the teacher. $A_{sup}$ monitors the classroom, obtains the phase transition signal, and decides whether to proceed to the next teaching phase or end the lesson. This can be represented as:
\begin{equation}
SIG(t) = A_{sup}(TP+TS+STA(t))
\end{equation}

With the help of $A_{sup}$, teachers can teach more effectively, and the interaction between teachers and students is more targeted, without deviating from the topic. During the questioning session, the supervisory Agent selects the most suitable student to ask questions based on the student's cognitive analysis of their willingness to speak. The supervisory Agent also monitors the persona status of the intelligent agents in real-time and maintains it if there's any deviation. Users can also operate the supervisory Agent to adjust the classroom process according to their needs.

\section{Experiments}
\label{sec:exp}

In this section, we first present the "classroom teaching scenario" reconstructed using the CGMI framework and analyze the teaching behaviors during the class. Subsequently, through comparative experiments, we showcase the behavioral advantages of agents equipped with human intrinsic traits (such as personality, cognitive structures, etc.). Lastly, we analyze the significance of generic intelligent agents in enhancing the interaction logic of role-specific agents.
In our experiment, we adopted OpenAI's gpt-3.5-turbo-16k model \cite{gpt3.5}, instantiating one teacher, five students, and four generic intelligent agents. Each agent was given a unique role setting and task objective (see appendix).

\subsection{Analysis of Teaching Behavior}
We employed the Flanders Interaction Analysis System (FIAS) to examine interactive behaviors between teachers and students across three virtual classroom sessions. We hired 2 trained experts to encode the teaching behaviors. These two encoders worked independently, encoding each sentence once and sequentially constructing a behavior sequence, ultimately achieving consistent evaluation results. These sessions focused on the following topics: C1: Concept of the Quadratic Equation, C2: Methods for Solving the Quadratic Equation, and C3: Applications of the Quadratic Equation. 

\begin{table}[t]

    \centering
    \begin{tabular}{l|lll}
    \hline
        Categories & C1 & C2 & C3 \\ \hline
        B1.Accept feeling & 0.35\% & 0\% & 0.30\% \\ 
        B2.Praises or encourages & 19.08\% & 12.99\% & 11.98\% \\ 
        B3.Accept ideas & 3.89\% & 6.39\% & 5.69\% \\ 
        B4.Asks questions & 1.77\% & 1.03\% & 1.50\% \\ 
    \hline
        B5.Lecturing & 22.97\% & 33.61\% & 35.61\% \\ 
        B6.Gives directions & 6.36\% & 7.01\% & 5.09\% \\ 
        B7.Criticising & 5.65\% & 1.24\% & 1.20\% \\ 
    \hline
        B8.Pupil talk response & 28.62\% & 20.41\% & 21.56\% \\ 
        B9.Pupil talk Initiation & 11.31\% & 17.32\% & 17.07\% \\ \hline
    \end{tabular}
        \caption{Analysis results based on FIAS}
    \label{tabel1} 
\end{table}


\begin{figure*}[t]
\centering
\includegraphics[width=0.9\linewidth]{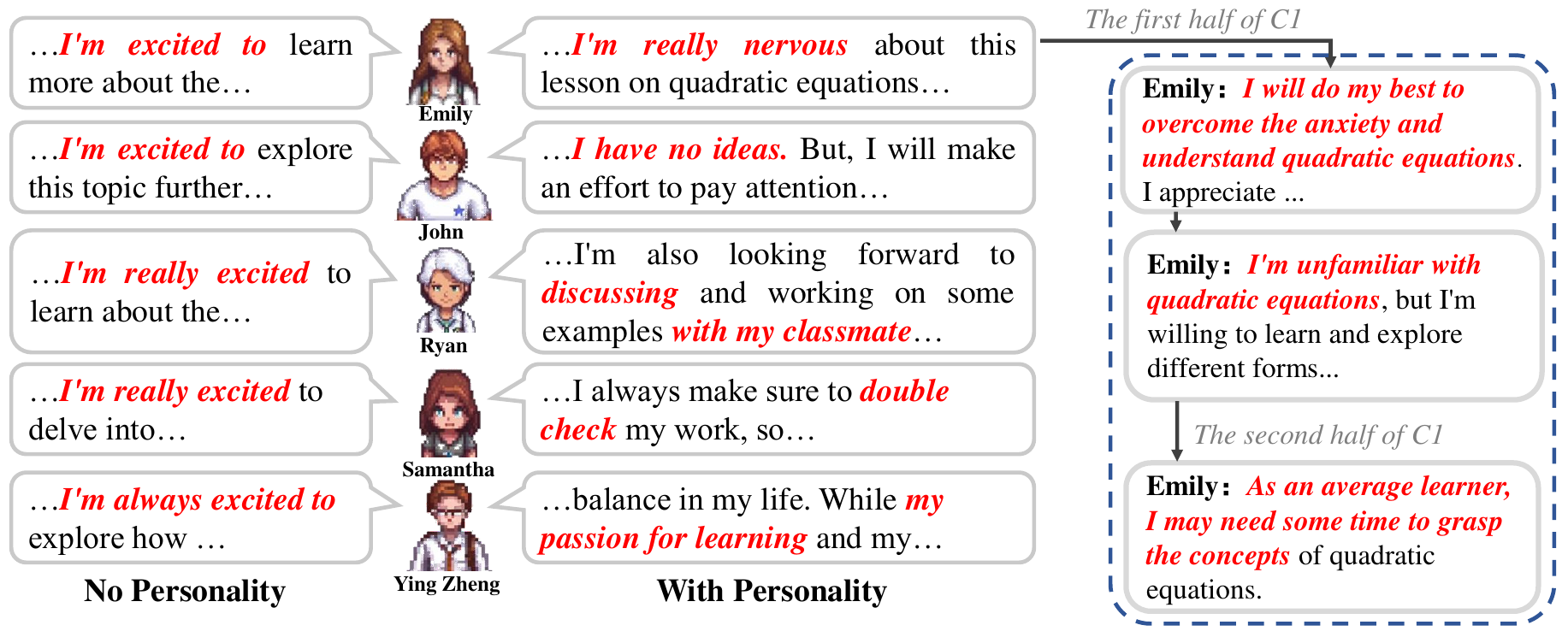}
\caption{The influence of personal traits on agent expression.} 
\label{fig5}
\end{figure*}

Table \ref{tabel1} shows the proportion of each interaction behavior in the course. Overall, the variety of interactions in the virtual classroom is rich and consistent with actual teaching, validating the effectiveness of CGMI by demonstrating its ability to effectively organize interactions and collaboration between multi-agents.

According to the results in table \ref{tabel1}, teacher's behavior(B1, B2, B3, B4, B5, B6, B7) made up an average of 61.23\% of the discourse in these mathematics sessions. In contrast, students' behavior(B8, B9) facilitated by teacher prompts represented an average of 23.53\%. Notably, the ratio of indirect influence behaviors (B1, B2, B3, B4) to direct influence behaviors (B5, B6, B7) remained below 1. This suggests that the virtual classroom is dominated by teachers who have direct control over the overall classroom. Furthermore, student-initiated interactions constituted about 15.23\%, suggesting that students remain engaged, deliberating, and responding to queries under the teacher's guidance.

\subsection{Intrinsic Characteristics of Intelligent Agents}

To assess the efficacy of the proposed cognitive architecture, we examined it through the lens of a teacher, Mrs. Smith, analyzing her classroom practices and her subsequent reflections and plans. As illustrated in Figure 4, we displayed the part of her reflective and planning processes within a single lesson and across two different lessons. Our analysis sought to elucidate the influence of the cognitive structure on agents, emphasizing the model's capacity for both reflection and planning. We analyzed the effectiveness of the algorithm from within and between classes.

\textbf{(1) Within the lesson}: In Course-ONE, student Emily conveyed her anxiety, stating, "{\textit{I'm really nervous about this lesson.}}" Mrs. Smith, attuned to this feedback, incorporated it into her reflective process and instructional planning. Drawing from a library of teaching techniques, she employed strategies such as heightened encouragement and gamified instructional methods. A parallel observation was made in Course-TWO and Course-THREE. Mrs. Smith prompted students to consider, “{\textit{How do coefficients 'b' and 'c' affect the graph of a quadratic function?}}”, and reiterated the topic in her subsequent planning. Following the actions of encouragement, Mrs. Smith's reflective records recognized her efforts in affirming and uplifting students.

\textbf{(2) Between lessons}: Across different courses, the proposed cognitive structure is still valid. It plays a crucial role in refining Mrs. Smith's teaching focus, deepening understanding and adapting teaching methods. For example, through reflection on Course-ONE, Mrs Smith found that Emily exhibited anxiety when faced with mathematical challenges. This insight directly influenced Mrs.Smith reassuring statement to Emily in Course-TWO: "\textit{I'm pleased to see you've overcome your apprehension towards mathematics.}"

\textbf{The effect of tree-structured persona model.} To discern whether agents with varied personality traits exhibit distinguishable behaviors during interactions, we executed a comparative study depicted in Figure 5. One lesson involved personality allocation, detection, and maintenance, whereas the other lacked any defined agent personalities. In the absence of assigned traits, there was a notable uniformity in the expressions of five students, often resorting to statements like, "\textit{I'm excited...}". In contrast, once unique personality traits were allocated, their expressions became more nuanced and aligned with their respective personas. For instance, the outgoing Ryan would suggest a “\textit{discussion with classmates}”, while the industrious Ying Zheng would exude a “\textit{passion for learning}”.

Furthermore, on the right side of Figure 5, the statements made by the student Emily throughout the class are displayed. Judging from the records of her remarks, the Emily Agent has demonstrated a consistent persona, interacting with teachers and classmates based on the previously established persona. In detail, she remarked, “\textit{I'm considerably anxious about this quadratic equations segment.}” at the start of the class. In the middle part of the course, she still showed her unfamiliarity and lack of confidence in the current knowledge in the interaction, expressing like, "\textit{I'm not well-versed with quadratic equations, yet I'm keen on learning and exploring various aspects...}", and “\textit{Being an average student, I might require a while to fully comprehend quadratic equations}”.

By imbuing agents with human-like qualities, they can adeptly distill insights from evolving scenarios and exhibit individualized responses. In addition, it also can make agents recalibrate actions based on accumulated knowledge and abilities. This significantly augments agents' adaptive capabilities in multifaceted environments. Concurrently, the tree-structured character model introduced in this study effectively and efficiently captures and retains the personalized data of agents.

\subsection{Quantitative Analysis of Interaction Logic}

\begin{figure}
\includegraphics[width=\linewidth]{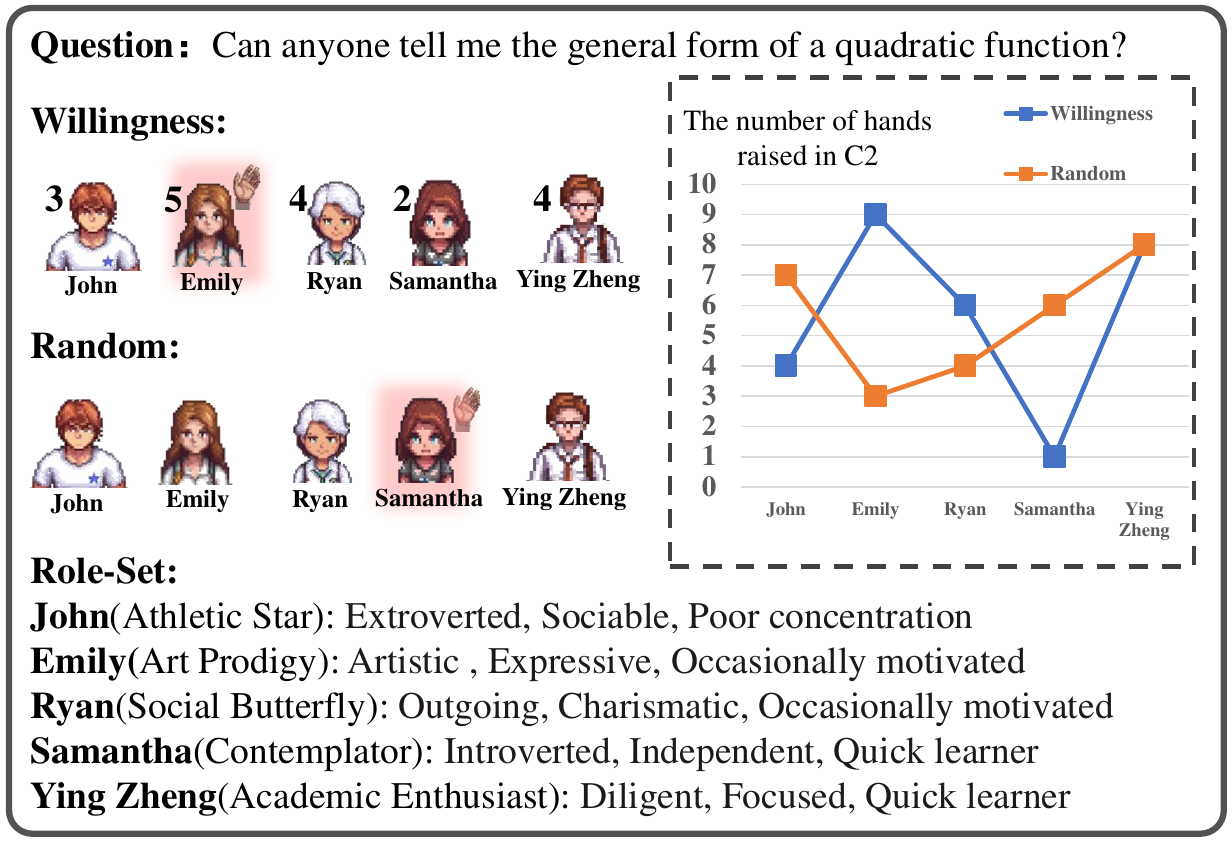}
\caption{The influence of personal traits on agent expression.} 
\label{fig6}
\end{figure}


Based on the "classroom teaching" scenario restored by CGMI, this paper compares the rationality of different interaction logics under the same question. 

\textbf{Analysis of willingness to speak.} As shown in the Figure \ref{fig6}, when the teacher posed the question to all students: "{\textit{Can anyone tell me the general form of a quadratic function?}", the outcomes differed between the answer willingness judgment agent and random selection methods. The former showed the students' willingness to answer intensity: John: 3, Emily: 5, Ryan: 4, Samantha: 2, Ying Zheng: 4. Notably, the students' willingness strength is highly consistent with their character traits. For instance, the expressive Emily exhibited high willingness to answer, while the introverted Samantha showed less. The random selection method, however, produced different results.

The discrepancy between the two methods is not coincidental. We recorded the number of students recommended by the two different methods to answer when the teacher posed questions to the entire class during a complete lesson. From the Figure \ref{fig6}, it can be seen that the answer willingness judgment agent, considering factors like students' personalities, classroom dynamics, and their grasp of the subject, recommended John 4 times, Emily 9 times, Ryan 6 times, Samantha 1 time, and Ying Zheng 8 times. However, with random selection, the results were John 7 times, Emily 3 times, Ryan 4 times, Samantha 6 times, and Ying Zheng 8 times. The expressive Emily only volunteered to answer 3 times, significantly undermining the rationality of the interaction process between the teacher and students in the virtual scenario.

\begin{figure}
\includegraphics[width=\linewidth]{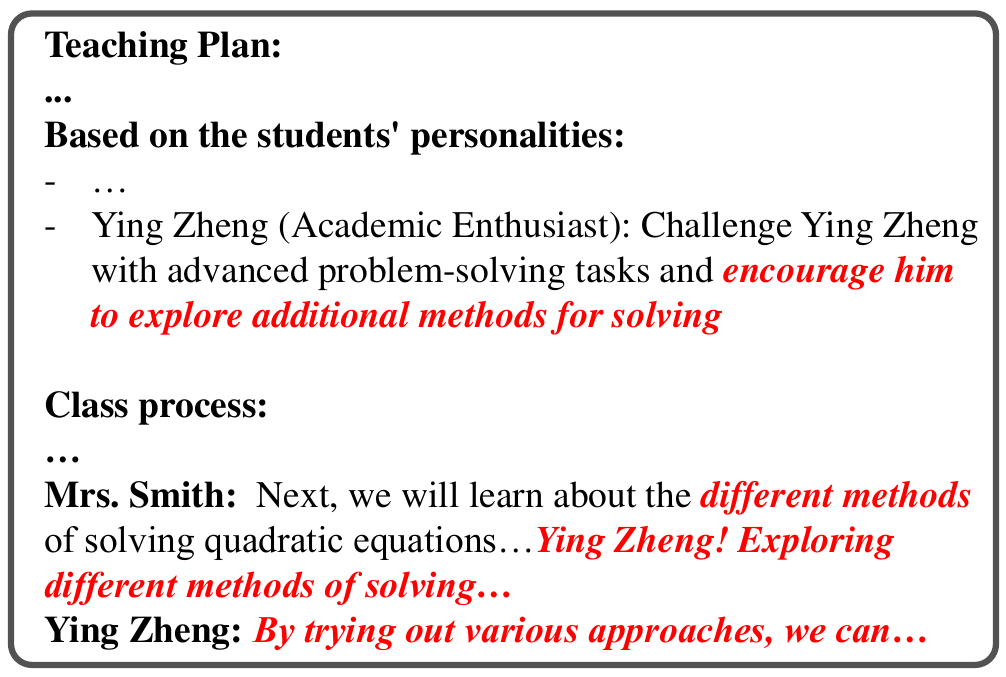}
\caption{The influence of personal traits on agent expression.} 
\label{fig8}
\end{figure}

\textbf{The effectiveness of questioning.} In addition to posing questions to all students, teachers also selectively direct questions to specific students. This selection is influenced by two aspects: (1) some teaching plans targeting particular students and (2) it's influenced by the teacher's analysis of the student's status and classroom dynamics during the teaching process. As shown in Figure \ref{fig8}, the teaching plan specifies that the teacher can encourage Ying Zheng to explore different solutions. As observed in the subsequent teaching process, the teacher aptly integrated this instructional arrangement during the lecture and specifically asked Ying Zheng to explore, leading to the next phase of instruction.

In summary, the flexible interaction logic setting ensures that the interaction process among multiple agents is no longer a random choice without considering the actual situation and role settings, nor a process where every role needs to be expressed. This introduces more possibilities for virtual scenarios.
\section{Conclusion}
\label{sec:con}

This paper introduces a multi-agent interaction framework (CGMI) that supports personalized configurations, enabling multiple agents to engage in anthropomorphic interactions and collaborations. It also can simulate domain-specific social phenomena. We designed a cognitive architecture equipped with domain skill library. It allows agents to combine domain knowledge for reflection and planning, and  condense the working memory into declarative and procedural memories. With the assistance of general agents, the authenticity of scenarios can be further enhanced. Moreover, we employed a virtual "classroom teaching" scenario to simulate the teaching process between teachers and students, and conducted comparative analysis of their interaction content and logic, verifying the effectiveness of CGMI. 

In the future, we hope that the social scenarios simulated by multiple agents will not only provide users with valuable social experimental data, aiding the development of large models, but also support industrial applications, such as assisting teaching and gamified teaching.

\bibliography{aaai24}

\begin{thebibliography}{27}
\providecommand{\natexlab}[1]{#1}

\bibitem[{Aher, Arriaga, and Kalai(2023)}]{pmlr-v202-aher23a}
Aher, G.~V.; Arriaga, R.~I.; and Kalai, A.~T. 2023.
\newblock Using Large Language Models to Simulate Multiple Humans and Replicate
  Human Subject Studies.
\newblock In Krause, A.; Brunskill, E.; Cho, K.; Engelhardt, B.; Sabato, S.;
  and Scarlett, J., eds., \emph{Proceedings of the 40th International
  Conference on Machine Learning}, volume 202 of \emph{Proceedings of Machine
  Learning Research}, 337--371. PMLR.

\bibitem[{Alexandru et~al.(2015)Alexandru, Tirziu, Tudora, and
  Bica}]{alexandru2015enhanced}
Alexandru, A.; Tirziu, E.; Tudora, E.; and Bica, O. 2015.
\newblock Enhanced education by using intelligent agents in multi-agent
  adaptive e-learning systems.
\newblock \emph{Studies in Informatics and Control}, 24(1): 13--22.

\bibitem[{Anderson and R(1983)}]{anderson1983spreading}
Anderson; and R, J. 1983.
\newblock A spreading activation theory of memory.
\newblock \emph{Journal of verbal learning and verbal behavior}, 22(3):
  261--295.

\bibitem[{Argyle et~al.(2023)Argyle, Busby, Fulda, Gubler, Rytting, and
  Wingate}]{argyle2023out}
Argyle, L.~P.; Busby, E.~C.; Fulda, N.; Gubler, J.~R.; Rytting, C.; and
  Wingate, D. 2023.
\newblock Out of one, many: Using language models to simulate human samples.
\newblock \emph{Political Analysis}, 31(3): 337--351.

\bibitem[{Bran et~al.(2023)Bran, Cox, White, and Schwaller}]{bran2023chemcrow}
Bran, A.~M.; Cox, S.; White, A.~D.; and Schwaller, P. 2023.
\newblock ChemCrow: Augmenting large-language models with chemistry tools.
\newblock arXiv:2304.05376.

\bibitem[{Davidsson and Paul(2002)}]{davidsson2002agent}
Davidsson; and Paul. 2002.
\newblock Agent based social simulation: A computer science view.
\newblock \emph{Journal of artificial societies and social simulation}, 5(1).

\bibitem[{Grigorenko and Sternberg(1993)}]{grigorenko1993thinking}
Grigorenko, E.; and Sternberg, R. 1993.
\newblock Thinking styles in teaching inventory.
\newblock \emph{unpublished test, Yale University}.

\bibitem[{Jiang et~al.(2023)Jiang, Zhang, Cao, and
  Kabbara}]{jiang2023personallm}
Jiang, H.; Zhang, X.; Cao, X.; and Kabbara, J. 2023.
\newblock PersonaLLM: Investigating the Ability of GPT-3.5 to Express
  Personality Traits and Gender Differences.
\newblock arXiv:2305.02547.

\bibitem[{John, Srivastava et~al.(1999)}]{john1999big}
John, O.~P.; Srivastava, S.; et~al. 1999.
\newblock The Big-Five trait taxonomy: History, measurement, and theoretical
  perspectives.

\bibitem[{Krishna et~al.(2022)Krishna, Lee, Fei-Fei, and
  Bernstein}]{krishna2022socially}
Krishna, R.; Lee, D.; Fei-Fei, L.; and Bernstein, M.~S. 2022.
\newblock Socially situated artificial intelligence enables learning from human
  interaction.
\newblock \emph{Proceedings of the National Academy of Sciences}, 119(39):
  e2115730119.

\bibitem[{Li et~al.(2023)Li, Hammoud, Itani, Khizbullin, and
  Ghanem}]{li2023camel}
Li, G.; Hammoud, H. A. A.~K.; Itani, H.; Khizbullin, D.; and Ghanem, B. 2023.
\newblock CAMEL: Communicative Agents for "Mind" Exploration of Large Scale
  Language Model Society.
\newblock arXiv:2303.17760.

\bibitem[{Mara~Pudane and Radin(2017)}]{PUDANE2017517}
Mara~Pudane, E.~L.; and Radin, M.~A. 2017.
\newblock Human Emotional Behavior Simulation in Intelligent Agents: Processes
  and Architecture.
\newblock \emph{Procedia Computer Science}, 104: 517--524.
\newblock ICTE 2016, Riga Technical University, Latvia.

\bibitem[{Markel et~al.(2023)Markel, Opferman, Landay, and
  Piech}]{markel2023gpteach}
Markel, J.~M.; Opferman, S.~G.; Landay, J.~A.; and Piech, C. 2023.
\newblock GPTeach: Interactive TA Training with GPT Based Students.

\bibitem[{Nair et~al.(2023)Nair, Schumacher, Tso, and Kannan}]{nair2023dera}
Nair, V.; Schumacher, E.; Tso, G.; and Kannan, A. 2023.
\newblock DERA: Enhancing Large Language Model Completions with Dialog-Enabled
  Resolving Agents.
\newblock arXiv:2303.17071.

\bibitem[{{OpenAI}(2022)}]{gpt3.5}
{OpenAI}. 2022.
\newblock OpenAI. Introducing chatgpt.
\newblock \url{https://openai.com/blog/chatgpt}.
\newblock Accessed: 2023-03-1.

\bibitem[{OpenAI(2023)}]{openai2023gpt4}
OpenAI. 2023.
\newblock GPT-4 Technical Report.
\newblock arXiv:2303.08774.

\bibitem[{Park et~al.(2023)Park, O'Brien, Cai, Morris, Liang, and
  Bernstein}]{park2023generative}
Park, J.~S.; O'Brien, J.~C.; Cai, C.~J.; Morris, M.~R.; Liang, P.; and
  Bernstein, M.~S. 2023.
\newblock Generative Agents: Interactive Simulacra of Human Behavior.
\newblock arXiv:2304.03442.

\bibitem[{Park et~al.(2022)Park, Popowski, Cai, Morris, Liang, and
  Bernstein}]{park2022social}
Park, J.~S.; Popowski, L.; Cai, C.; Morris, M.~R.; Liang, P.; and Bernstein,
  M.~S. 2022.
\newblock Social Simulacra: Creating Populated Prototypes for Social Computing
  Systems.
\newblock In \emph{Proceedings of the 35th Annual ACM Symposium on User
  Interface Software and Technology}, UIST '22. New York, NY, USA: Association
  for Computing Machinery.
\newblock ISBN 9781450393201.

\bibitem[{Press et~al.(2023)Press, Zhang, Min, Schmidt, Smith, and
  Lewis}]{press2023measuring}
Press, O.; Zhang, M.; Min, S.; Schmidt, L.; Smith, N.~A.; and Lewis, M. 2023.
\newblock Measuring and Narrowing the Compositionality Gap in Language Models.
\newblock arXiv:2210.03350.

\bibitem[{Qian et~al.(2023)Qian, Cong, Yang, Chen, Su, Xu, Liu, and
  Sun}]{qian2023communicative}
Qian, C.; Cong, X.; Yang, C.; Chen, W.; Su, Y.; Xu, J.; Liu, Z.; and Sun, M.
  2023.
\newblock Communicative Agents for Software Development.
\newblock arXiv:2307.07924.

\bibitem[{Qian et~al.(2018)Qian, Huang, Zhao, Xu, and Zhu}]{qian2018assigning}
Qian, Q.; Huang, M.; Zhao, H.; Xu, J.; and Zhu, X. 2018.
\newblock Assigning Personality/Profile to a Chatting Machine for Coherent
  Conversation Generation.
\newblock In \emph{Ijcai}, 4279--4285.

\bibitem[{Soloman and Felder(2005)}]{soloman2005index}
Soloman, B.~A.; and Felder, R.~M. 2005.
\newblock Index of learning styles questionnaire.
\newblock \emph{NC State University. Available online at: http://www. engr.
  ncsu. edu/learningstyles/ilsweb. html (last visited on 14.05. 2010)}, 70.

\bibitem[{Wang et~al.(2023{\natexlab{a}})Wang, Xu, Lan, Hu, Lan, Lee, and
  Lim}]{wang2023planandsolve}
Wang, L.; Xu, W.; Lan, Y.; Hu, Z.; Lan, Y.; Lee, R. K.-W.; and Lim, E.-P.
  2023{\natexlab{a}}.
\newblock Plan-and-Solve Prompting: Improving Zero-Shot Chain-of-Thought
  Reasoning by Large Language Models.
\newblock arXiv:2305.04091.

\bibitem[{Wang et~al.(2023{\natexlab{b}})Wang, Cai, Liu, Ma, and
  Liang}]{wang2023describe}
Wang, Z.; Cai, S.; Liu, A.; Ma, X.; and Liang, Y. 2023{\natexlab{b}}.
\newblock Describe, Explain, Plan and Select: Interactive Planning with Large
  Language Models Enables Open-World Multi-Task Agents.
\newblock arXiv:2302.01560.

\bibitem[{Wang et~al.(2023{\natexlab{c}})Wang, Mao, Wu, Ge, Wei, and
  Ji}]{wang2023unleashing}
Wang, Z.; Mao, S.; Wu, W.; Ge, T.; Wei, F.; and Ji, H. 2023{\natexlab{c}}.
\newblock Unleashing Cognitive Synergy in Large Language Models: A Task-Solving
  Agent through Multi-Persona Self-Collaboration.
\newblock arXiv:2307.05300.

\bibitem[{{Weng and Lilian}(2023)}]{weng2023prompt}
{Weng and Lilian}. 2023.
\newblock LLM-powered Autonomous Agents.
\newblock \url{https://lilianweng.github.io/posts/2023-06-23-agent/}.
\newblock Accessed: 2023-06-23.

\bibitem[{Yao et~al.(2023)Yao, Zhao, Yu, Du, Shafran, Narasimhan, and
  Cao}]{yao2023react}
Yao, S.; Zhao, J.; Yu, D.; Du, N.; Shafran, I.; Narasimhan, K.; and Cao, Y.
  2023.
\newblock ReAct: Synergizing Reasoning and Acting in Language Models.
\newblock arXiv:2210.03629.

\end{thebibliography}

\clearpage
\section{Appendix}
\label{sec:app}
The appendix presents the character settings for each character, a tree-structured learning style scale, and a teaching style scale.

\subsection{Role Set}
In this work, the initialization of role agents is mainly carried out from the perspectives of the career, name, basic information, personalities, and teaching or learning styles. Figure 8 shows Teacher Mrs Smith's character settings. Figures 9, 10, 11, 12, and 13 show the character settings of students Ryan, John, Emily, Samantha, and Ying Zheng, respectively.

\subsection{Sternberg Thinking Styles in Teaching}
Mrs. Smith's teaching style can be described by Sternberg Thinking Styles in Teaching Inventory with a tree-structured format (Figure 14). Each Level-2 node has its score, representing the degree of match between the description provided and the actual teaching style, with a maximum of 7 and a minimum of 1. Each Level-1 node also has its corresponding score, which is the sum of the scores of all its child nodes. The higher the value, the higher the degree of matching.

\subsection{Solomon's Learning Styles}
Students learning styles can be described by Solomon's Learning Styles Inventory with a tree-structured format (Figure 15). Each Level-1 node has its type to represent your type in four different dimensions. When selecting 11 sub-nodes, a is selected more times than b, then the category represented is the former in the description, otherwise, it is the latter. Each Level-2 node has its description and choice to indicate your selection for the current evaluation question. 

\begin{figure}[htb]
\includegraphics[width=\linewidth]{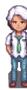} 
\caption{Character setting for Mrs. Smith.}
\label{app-fig1}
\end{figure}

\begin{figure}[htb]
\includegraphics[width=\linewidth]{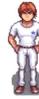} 
\caption{Character setting for Ryan.}
\label{app-fig4}
\end{figure}
\vspace{-0.4cm}

\begin{figure}[htb]
\includegraphics[width=\linewidth]{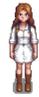} 
\caption{Character setting for John.}
\label{app-fig2}
\end{figure}
\vspace{-0.4cm}

\begin{figure}[htb]
\includegraphics[width=\linewidth]{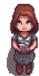} 
\caption{Character setting for Emily.}
\label{app-fig3}
\end{figure}
\vspace{-0.4cm}

\begin{figure}[htb]
\includegraphics[width=\linewidth]{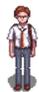} 
\caption{Character setting for Samantha.}
\label{app-fig5}
\end{figure}
\vspace{-0.4cm}

\begin{figure}[htb]
\includegraphics[width=\linewidth]{fig/Ying_Zheng.pdf} 
\caption{Character setting for Mrs. Smith.}
\label{app-fig6}
\end{figure}

\begin{figure*}[b]
\includegraphics[width=\linewidth]{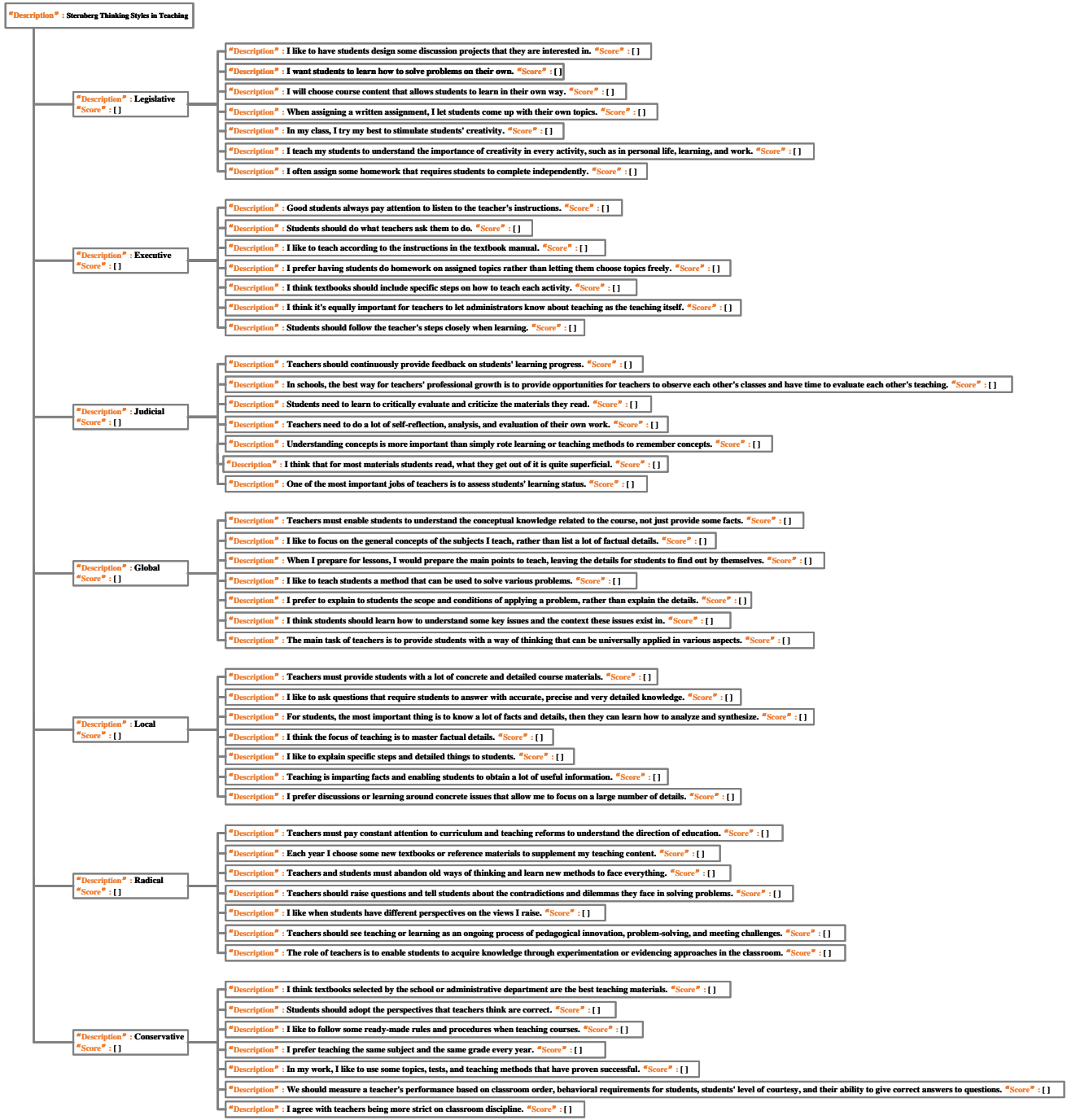} 
\caption{The Sternberg Thinking Styles in Teaching Inventory.}
\label{app-fig7}
\end{figure*}

\begin{figure*}[b]
\includegraphics[width=\linewidth]{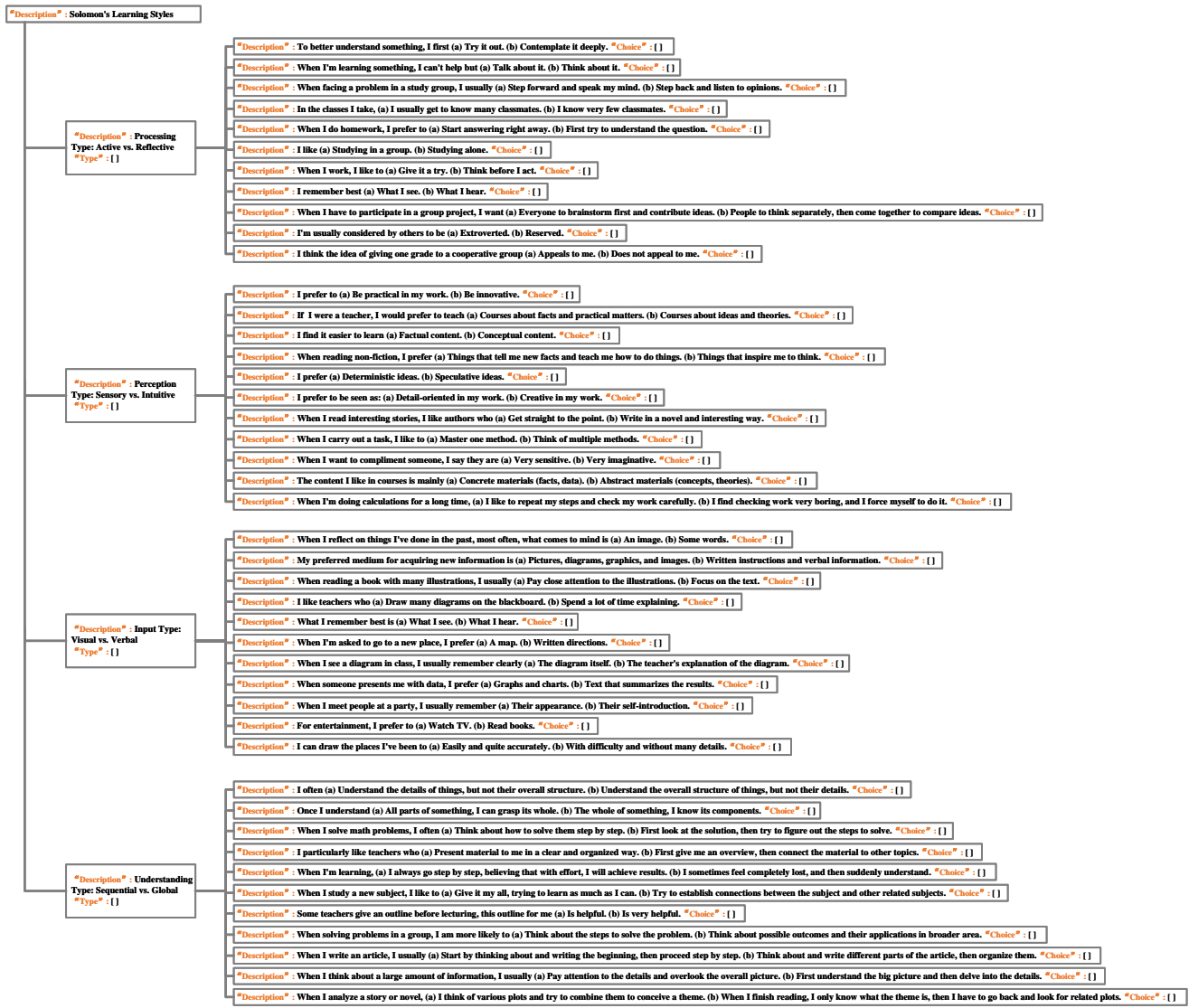} 
\caption{The Solomon's Learning Styles Inventory.}
\label{app-fig8}
\end{figure*}

\end{document}